\documentclass{article}

\usepackage{hyperref}
\usepackage{url}
\usepackage{graphicx} 
\usepackage{pgfplots}
\usepackage{pgfplotstable}
\usepgfplotslibrary{groupplots}
\pgfplotsset{compat=1.18}
\usepackage{booktabs}
\usepackage{multirow}

\newcommand{\mstd}[2]{\ensuremath{#1_{\scriptscriptstyle #2}}}

\usepackage{makecell}
\usepackage{booktabs} 
\usepackage{siunitx}
\usepackage{enumitem}
\usepackage{tabularx}
\newcommand{\bbit}{\texttt{32b}}
\newcommand{\ebit}{\texttt{8b}}
\newcommand{\dyn}{\textsc{D}} 
\newcommand{\linq}{\textsc{L}} 
\usepackage[dvipsnames]{xcolor}
\usepackage{xcolor}
\usepackage{wrapfig}
\usetikzlibrary{patterns}
\usepgfplotslibrary{groupplots}
\usetikzlibrary{calc}
\usepackage[table]{xcolor}
\usepgfplotslibrary{groupplots,fillbetween}
\definecolor{adamA}{RGB}{33,114,255}     
\definecolor{adamB}{RGB}{120,170,255}    
\definecolor{muonA}{RGB}{0,150,110}      
\definecolor{muonB}{RGB}{110,195,165}    
\definecolor{hybA}{RGB}{155,80,200}      
\definecolor{hybB}{RGB}{200,150,230}     

\definecolor{adamRed}        {RGB}{186,60,54}
\definecolor{adamRedDark}    {RGB}{156,50,47}
\definecolor{adamRedLight}   {RGB}{232,172,164}

\definecolor{muonBlue}       {RGB}{52,98,152}
\definecolor{muonBlueDark}   {RGB}{44,83,128}
\definecolor{muonBlueLight}  {RGB}{171,189,214}

\definecolor{fullGreen}      {RGB}{88,129,87}
\definecolor{fullGreenDark}  {RGB}{72,108,71}

\definecolor{dpo}{HTML}{D46A6A}   
\definecolor{simpo}{HTML}{6A8FD4} 

\definecolor{aRedDark}{RGB}{170,54,58}
\definecolor{aRed}{RGB}{230,102,106}
\definecolor{aRedLight}{RGB}{247,177,179}

\definecolor{mBlueDark}{RGB}{64,90,160}
\definecolor{mBlue}{RGB}{106,142,220}
\definecolor{mBlueLight}{RGB}{175,196,238}

\definecolor{fGreenDark}{RGB}{82,122,82}
\definecolor{fGreen}{RGB}{124,178,124}

\definecolor{adam}{HTML}{BF4E45}  
\definecolor{muon}{HTML}{3B6FB6}  
\definecolor{mgreen}{HTML}{4C9A6A} 

\definecolor{bestcol}{gray}{0.9}

\definecolor{MuonFamily}{HTML}{0072B2}     
\definecolor{FullMuonFamily}{HTML}{D55E00} 
\newif\ifshowcomments
\showcommentstrue 

\usepackage[preprint]{icml2026}

\usepackage[textsize=tiny]{todonotes}

\icmltitlerunning{Effective Quantization of Muon Optimizer States}

\input{Definitions}

\begin{document}

\twocolumn[
  \icmltitle{Effective Quantization of Muon Optimizer States}

  \icmlsetsymbol{equal}{*}

  \begin{icmlauthorlist}

    \icmlauthor{Aman Gupta}{equal,Nubank}
    \icmlauthor{Rafael Celente}{equal,Nubank}
    \icmlauthor{Abhishek Shivanna}{equal,Nubank}
    \icmlauthor{D. T. Braithwaite}{equal,Nubank}
    \icmlauthor{Gregory Dexter$^\dagger$}{equal,Nubank}
    \icmlauthor{Shao Tang$^\dagger$}{equal,Nubank}
    \icmlauthor{Hiroto Udagawa}{Nubank}
    \icmlauthor{Daniel Silva}{Nubank}
    \icmlauthor{Rohan Ramanath}{Nubank}
    \icmlauthor{S. Sathiya Keerthi}{Nubank}

  \end{icmlauthorlist}

  \icmlaffiliation{Nubank}{Nubank}

  \icmlcorrespondingauthor{Aman Gupta}{\texttt{aman.gupta@nubank.com.br}}
  \icmlcorrespondingauthor{Gregory Dexter}{\texttt{gregory.dexter@nubank.com.br}}
  \icmlcorrespondingauthor{Shao Tang}{\texttt{tang.shao@nubank.com.br}}
  \icmlkeywords{Machine Learning, Optimizer Quantization, Muon Optimizer}    

  \vskip 0.3in
]
\printAffiliationsAndNotice{\icmlEqualContribution \\ $^\dagger$Work done while at LinkedIn.}




\begin{abstract}
The Muon optimizer, based on matrix orthogonalization, has recently shown faster convergence and better computational efficiency over AdamW in LLM pre-training. However, the memory overhead of maintaining high-precision optimizer states remains a challenge for large-scale deployment. In this paper, we introduce the 8-bit Muon optimizer using blockwise quantization. 

In extensive Chinchilla-optimal experiments on pre-training models of up to 2.7B in size and fine-tuning them for instruction following, we demonstrate that 8-bit Muon achieves parity with Muon in terms of validation loss and downstream benchmarks, while achieving up to a 62\% reduction in optimizer state footprint. Crucially, we show that Muon's update mechanism is uniquely compatible with a simple linear quantization scheme, bypassing the complex dynamic scaling required for quantized AdamW. We supplement our empirical findings with a theoretical analysis of Muon's robustness to quantization noise.

\end{abstract}

\section{Introduction}

Scaling laws for large language models (LLMs)~\citep{DBLP:journals/corr/abs-2001-08361,hoffmann2022training} indicate that larger models generally achieve better out-of-distribution performance across diverse tasks. Yet, GPU high-bandwidth memory (HBM) capacity has not kept pace with parameter counts. During training, memory is dominated by model parameters, gradients, optimizer states, and activations. Systems work has therefore focused on distributing these tensors across devices via distributed data parallel (DDP), Fully Sharded Data Parallel (FSDP)~\citep{zhao2023pytorch}, ZeRO stage-3 in DeepSpeed~\citep{rajbhandari2020zero}, and tensor/model parallelism~\citep{shoeybi2019megatron} in order to improve training performance.

Orthogonal to sharding is compressing the optimizer state. AdamW~\citep{DBLP:journals/corr/abs-1711-05101,kingma2014adam}, the \emph{de facto} optimizer for LLMs, maintains two FP32 moment buffers (first and second moments) per parameter. For an 8B-parameter model (e.g., an 8B Llama-3 variant~\citep{dubey2024llama}), this alone occupies $64$ GB ($\sim$80\% of an NVIDIA H100’s 80\,GB HBM), leaving little headroom for parameters, gradients, and activations. To mitigate this,~\citet{dettmers20218} quantize Adam’s optimizer states to 8 bits via block-wise \emph{dynamic} (non-linear) quantization, preserving stability in the presence of extreme values while reducing optimizer memory by roughly $4\times$, enabling performant training under tight memory budgets.

Recently, there has been a surge of interest in moving beyond AdamW to improve training efficiency~\citep{anil2020scalable, shazeer2018adafactor, vyas2024soap}. Among various advances, one particularly promising optimizer is \textbf{Muon}~\citep{jordan2024muon}\footnote{Muon is closely related to SGD with momentum, adding a per-layer matrix orthogonalization}, which orthogonalizes the gradient momentum before updating the parameters. Equalizing the importance of all update directions results in improved stability and better convergence~\citep{bernstein2024modular, bernstein2025derivingmuon}. Several large-scale studies have confirmed Muon's ability to achieve improved convergence rate to a target validation loss compared to AdamW on a compute-optimal setup ~\citep{liu2025muon, shah2025practical}. Muon has also been used to train extremely large models up to a trillion parameters in size, like Kimi K2~\citep{team2025kimi} and GLM4.5~\citep{zeng2025glm}. While Muon offers improved convergence over AdamW, it still maintains FP32 optimizer state that competes for scarce HBM alongside parameters, gradients, and activations. As models continue to scale, this memory pressure remains a practical constraint even when the underlying optimizer improves.

\subsection{Contributions}

In this paper, we introduce the \textbf{8-bit Muon} optimizer with blockwise quantization. Whereas 8-bit AdamW variants are usually stable only with dynamic quantization, we demonstrate that \textbf{8-bit Muon can handle both linear and dynamic quantization effectively}.

We systematically study approaches to quantizing the Muon optimizer by pretraining GPT models of up to 2.7B parameters with Chinchilla-optimal data ratios on the FineWeb-Edu set. Notably, we show that dynamic and linear quantization of Muon optimizer updates for pre-training workloads demonstrate parity in terms of quality - both for validation loss and downstream benchmarks - while achieving a \emph{62\%} reduction in memory usage at the 2.7B model size. This memory reduction can be improved to 75\% by additionally applying dynamic quantization to non-hidden-layer parameters updated by AdamW, at the cost of up to a 0.5\% increase in validation loss. We observe similar results under Supervised Fine-Tuning (SFT), where memory efficiency is especially important in memory-constrained deployment settings. 

A key empirical finding is that \emph{linear} quantization of Muon optimizer state achieves parity with dynamic quantization and full-precision Muon in training progress. This contrasts with AdamW, where even sophisticated dynamic quantization does not fully retain performance, and linear quantization results in model divergence. This motivates our mechanistic analysis below. 

Prior work by \citet{dettmers20218} observed that naive 8-bit linear quantization can make AdamW unstable, and motivated block-wise dynamic quantization by highlighting sensitivity to quantization error in the optimizer states, especially the second-moment buffer. In Theorem \ref{thm:adam_quant}, we formalize this mechanism by showing that AdamW can exhibit unbounded error under linear quantization, and we identify the second-moment accumulator in the denominator as the critical driver. This points to a natural control case: SGD with momentum, which removes the second-moment term. In Theorem \ref{thm:sgd_quant}, we show that linearly-quantized SGD+M admits bounded error at each optimizer step, and empirically verify that it remains effective in practice. Finally, in Theorem \ref{thm:muon_quant_true_polar} we extend the same perspective to Muon, proving an analogous error bound with an additional dependence on the spectrum of the momentum matrix.

\textbf{Key takeaway:} Linear 8-bit quantization provides a simple and practical default to compress Muon’s optimizer state, achieving quality parity while substantially reducing optimizer state memory.


\begin{table*}[t]
\centering
\small
\begin{tabularx}{\linewidth}{@{}l c c l@{}}
\toprule
\textbf{Model} & \textbf{Muon state} & \textbf{AdamW state} & \textbf{Shorthand} \\
\midrule
32-bit AdamW & --- & \bbit & \texttt{AdamW-32} \\
32-bit Muon  & \bbit & \bbit & \texttt{Muon-32} \\
8-bit AdamW (dynamic) & --- & \ebit\ \dyn & \texttt{AdamW-8D} \\

8-bit Muon (linear), 8-bit AdamW (dynamic) & \ebit\ \linq & \ebit\ \dyn & \texttt{Muon-8L/AdamW-8D} \\
8-bit Muon (linear), 32-bit AdamW  & \ebit\ \linq & \bbit & \texttt{Muon-8L/AdamW-32} \\
8-bit Muon (dynamic), 8-bit AdamW (dynamic) & \ebit\ \dyn & \ebit\ \dyn & \texttt{Muon-8D/AdamW-8D} \\
8-bit Muon (dynamic), 32-bit AdamW & \ebit\ \dyn & \bbit & \texttt{Muon-8D/AdamW-32} \\
\bottomrule
\end{tabularx}
\caption{Optimizer variants considered for pre-training and SFT. \textsc{D} and \textsc{L} denote dynamic and linear quantization, respectively. The 8-bit Muon family comprises five studied variants. AdamW-8L related variants are unstable under Theorem~\ref{thm:adam_quant}, consistent with empirical observations, and hence omitted from the table. 
}
\label{tab:optimizer-variants}
\end{table*}

\section{Related Work}

\textbf{Efficient Optimizers} There has been some prior work on alleviating the memory cost of training with optimizers like AdamW. Techniques like low-rank adaptation (LoRA)~\citep{hu2022lora} allow a small subset of parameters to be fine-tuned for downstream tasks, but often fall behind in quality when compared to full parameter fine-tuning~\citep{biderman2024lora}. Adafactor~\citep{shazeer2018adafactor} factorizes the second moment for matrices, reducing memory consumption compared to AdamW. ~\citet{dettmers20218} introduced the 8-bit Adam optimizer, but it only works when combined with careful blockwise + dynamic quantization. Adam GaLore~\citep{zhao2024galore} leverages the low-rank structure of the gradients to reduce state size and can be combined with state quantization. Our work is directly comparable to \citet{dettmers20218}'s work, and can potentially be combined with low-rank updates.

\textbf{Gradient Orthogonalization} The Muon optimizer~\citep{jordan2024muon} has sparked interest in algorithms that take advantage of the orthonormalization of the gradient of matrix-valued parameters. Muon makes each weight matrix update orthonormal through polar decomposition (Newton-Schulz), giving direction-only, spectrally controlled steps for hidden layers. To scale it to large LLMs, recent works add decoupled weight decay and careful per-parameter update scaling~\citep{liu2025muon}. 
Dion~\citep{ahn2025dion} leverages orthonormalization but is built for distributed training: using low-rank orthonormalization with device-local momentum/error-feedback to avoid reconstruction or synchronization of full matrices.

\textbf{Quantization} Quantization is a versatile tool for managing the memory cost of training large models. While we apply quantization to the state of the Muon optimizer, it has also been successfully applied to model weights in two settings - post-training quantization (PTQ) and quantization-aware training (QAT). PTQ uses calibration data to quantize the weights of large models in one shot to $k$ bits, where $k$ can be as low as 1 or 2~\citep{tseng2024quip, frantar2022gptq, lin2024awq, behdin2023quantease}. QAT involves training with quantized weights~\citep{liu2023llm}. Both PTQ and QAT require hardware support to realize the full benefits of quantization. Another interesting work is MuLoCo~\citep{therien2025muloco}, where the authors apply Muon as the inner (local) optimizer in a DiLoCo-style~\citep{douillard2311diloco} loop, geared toward compressing parameter updates during distributed training. Thus, MuLoCo/DiLoco use quantization only for gradient updates.

\vspace{-3mm}
\section{Background}
\subsection{Quantization for Optimizers}

Quantization is the process of reducing the precision of numerical representations by mapping a value in a larger set to one in a smaller discrete set. For example, representing a real number as a 32-bit floating-point value, or converting a 32-bit float into an 8-bit integer, are both forms of quantization. In deep learning, model parameters and optimizer states are typically stored as 32-bit floating-point numbers, making their quantization to lower-precision formats a primary goal for memory efficiency.

 For instance, linear quantization is a method  (among many) to quantize the state tensor $\xb$ of an optimizer from 32-bit floats to 8-bit integers. Specifically, the $8$-bit integer linear quantization of $\zb$ may be computed as:
 \begin{gather*}
     \zb = \operatorname{round}\Big(\frac{127}{\max_i |\xb_i|} \cdot \xb\Big),
 \end{gather*}
 where $\round$ denotes rounding to the nearest integer entry-wise. Note that all entries of $\xb$ are integers in the range $-127$ to $127$ and hence may be exactly represented by an $8$-bit signed integer. This state $\zb$ may then be de-quantized by $\xbtil = (\max_i |\xb_i| / 127) \cdot \zb$, and maintains the property that $|\xbtil_j - \xb_j| \leq \|\xb\|_\infty / 254$ for all $j$, guaranteeing bounded reconstruction error.
 
For 8-bit quantization, the number of addressable states is 256. ~\citet{dettmers20218} discuss another quantization codebook design - dynamic quantization. Dynamic quantization is a more sophisticated approach that quantizes non-uniformly by allocating more codes to regions with high densities and fewer codes to sparsely used regions. This increases resilience to non-uniform data.

A crucial recipe for reducing the effect of outliers is blockwise quantization~\citep{dettmers20218}, where one can segment the optimizer state into one or more blocks, and then perform quantization (linear, dynamic, or other schemes) separately within each block. A nice side effect of blockwise quantization is that each block can be processed in parallel. For a more detailed treatment of blockwise and dynamic quantization, please refer to~\citep{dettmers20218}. 

\subsection{The Muon Algorithm}

The Muon update on a single hidden layer can be described as follows:

\begin{equation}
\begin{aligned}
\mathbf{M}^{(t)} &:= \beta\,\mathbf{M}^{(t-1)} + \nabla f\!\big(\mathbf{W}^{(t-1)}\big),\\
\mathbf{O}^{(t)} &:= \mathrm{NS}\!\big(\mathbf{M}^{(t)}\big),\\
\mathbf{W}^{(t)} &:= \mathbf{W}^{(t-1)} - \alpha\,\mathbf{O}^{(t)}.
\end{aligned}
\label{eq:muon-1}
\end{equation}

where $t$ is the iteration number and $\mathbf{M}$ is the momentum of the gradient. NS stands for the Newton-Schulz iteration process~\citep{bernstein2024old, higham2008functions}, used to find an approximation for the polar factor $\mathbf{U} \mathbf{V}^T$ where $\mathbf{U} \mathbf{\Sigma} \mathbf{V}^T$ is the singular value decomposition (SVD) of $\mathbf{M}$. Orthogonalization equalizes the importance of each update direction by collapsing all singular values to 1.

The vanilla version of Muon described above does not use weight decay. Additionally, it is not obvious whether it requires any hyperparameter tuning over AdamW baselines. \citet{liu2025muon} introduced a variant of Muon that uses weight decay and also scales its update to match the update RMS of AdamW, producing the following version:
\begin{equation}
\begin{aligned}
\mathbf{M}^{(t)} &:= \beta\,\mathbf{M}^{(t-1)} + \nabla f\!\big(\mathbf{W}^{(t-1)}\big),\\
\mathbf{O}^{(t)} &:= \mathrm{NS}\!\big(\mathbf{M}^{(t)}\big),\\
\mathbf{W}^{(t)} &:= (1-\alpha\lambda)\mathbf{W}^{(t-1)} - 0.2\alpha\sqrt{\max(m,n)}\mathbf{O}^{(t)}.
\end{aligned}
\label{eq:muon-2}
\end{equation}

where $m$ and $n$ are the dimensions of $\mathbf{M}$. \citet{liu2025muon} claim that with eqn.~(\ref{eq:muon-2}), hyperparameters such as learning rate and weight decay can be shared across matrix and non-matrix parameters. In the rest of the paper, any mention of Muon refers to the version in eqn.~(\ref{eq:muon-2}), unless stated otherwise. It is important to note that any non-matrix parameters and input/output parameters are optimized using AdamW, leaving Muon to focus on matrix-valued hidden parameters.

\section{Experiments}

This section presents a thorough empirical study of using linear and/or dynamic quantization to improve the memory efficiency of the Muon optimizer with minimal quality degradation. Recall that Muon is applied only to the two-dimensional hidden-layer weight matrices, while AdamW is used for all remaining parameters; see Table~\ref{tab:optimizer-variants} for details.

\subsection{The Quantized Muon Algorithm}

Throughout our experiments, we use the Muon variant of \cite{liu2025muon}, while compressing the optimizer state stored between iterations via linear or dynamic blockwise 8-bit quantization.

\begin{equation}
\begin{aligned}
\widetilde{\mathbf{M}}^{(t-1)} &:= \operatorname{Dequantize}_\Scal\big(\mathbf{Z}^{(t-1)}\big),\\[2pt]
\mathbf{M}^{(t)} &:= \beta\,\widetilde{\mathbf{M}}^{(t-1)} + \nabla f\!\big(\mathbf{W}^{(t-1)}\big),\\[2pt]
\mathbf{O}^{(t)} &:= \mathrm{NS}\!\big(\mathbf{M}^{(t)}\big),\\[2pt]
\mathbf{W}^{(t)} &:= (1-\alpha\lambda)\,\mathbf{W}^{(t-1)}
                  \;-\; 0.2\alpha\sqrt{\max(m,n)}\mathbf{O}^{(t)},\\[4pt]
\mathbf{Z}^{(t)} &:= \operatorname{Quantize}_\Scal\big(\mathbf{M}^{(t)}\big).
\end{aligned}
\end{equation}

where $\mathbf{Z}$ denotes the compressed momentum buffer and $\mathcal{S}$ denotes the associated auxiliary state used for dequantization. The remaining notation follows Equation~\ref{eq:muon-2}.

\label{sxn:experiments_main}

\subsection{Pre-training with 8-bit Muon}

\textbf{Architectures.}
For pre-training, we train a modified GPT-2 architecture~\citep{radford2019language} from scratch, replacing learned absolute positional embeddings with rotary positional embeddings (RoPE)~\citep{su2024roformer}. To study scaling, we consider five model sizes: Small (162M), Medium (405M), Large (834M), XL (1.4B), and XXL (2.7B). For brevity, we refer to this architecture family as GPT throughout the paper. Additional architectural details are provided in Table~\ref{table:model_architectures} in Appendix~\ref{appendix:pretrain}.

\textbf{Datasets.}
We pre-train on FineWeb-Edu~\citep{penedo2024fineweb}. Following the compute-optimal scaling guidance of~\citet{hoffmann2022training}, we use approximately 20 tokens per parameter, yielding a training set size that increases with model size (the \emph{Chinchilla-optimal} setting). For the 2.7B model, this corresponds to roughly 54B FineWeb tokens. We evaluate validation loss on 150K samples from the FineWeb validation split, totaling approximately 300M tokens.

\textbf{Training details.}
We pre-train the GPT models from scratch on FineWeb. For AdamW, we use $\beta_1=0.9$, $\beta_2=0.95$, and $\epsilon=10^{-8}$. For Muon, we set the momentum parameter to $0.95$. A complete list of hyperparameters and training configurations can be found in Appendix~\ref{appendix:pretrain}. For quantized AdamW and Muon variants, we use a block size of 2048. 
We fix the decoupled weight decay to 0.1 in all experiments.

We adopt the warmup-stable-decay (WSD) learning rate schedule~\citep{hu2024minicpm}, with linear warmup from 0 to the peak learning rate over the first 10\% of training steps, and linear decay to zero over the final 10\% of steps. For each model size, we use the peak learning rate and global batch size recommended by~\citet{brown2020language}, which were tuned for GPT-style models trained using AdamW. For Muon, we do not tune the peak learning rate; instead, we reuse the AdamW peak learning rates, consistent with our goal that Muon should serve as a drop-in replacement for AdamW without requiring learning-rate or weight-decay re-tuning~\citep{liu2025muon}.

\textbf{Evaluation criteria} Our evaluation of pre-training is two-pronged (a) validation loss and (b) benchmark performance on six different tasks using the \textbf{lm-eval} harness~\citep{eval-harness}. These tasks include MMLU~\citep{hendrycks2020measuring}, LAMBADA~\citep{paperno2016lambada}, BoolQ~\citep{clark2019boolqexploringsurprisingdifficulty}, HellaSwag~\citep{zellers2019hellaswagmachinereallyfinish}, ARC-Challenge and ARC-Easy~\citep{clark2018thinksolvedquestionanswering}. 

\subsubsection{Results}

\begin{figure}[t]
  \centering
  \includegraphics[width=\columnwidth]{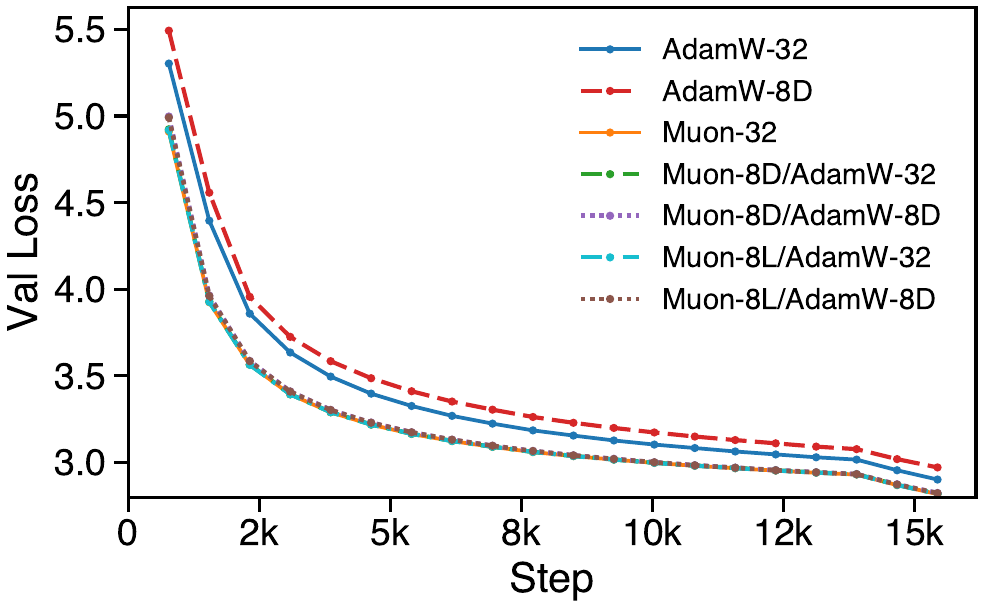}
  \caption{Validation loss for GPT-2 Medium across 7 variants.}
  \label{fig:gpt2-medium-demo}
\end{figure}


\paragraph{Medium-scale ablation: stability and sensitivity to quantization.}
We first perform a controlled ablation on GPT-Medium, sweeping all stable 8-bit Muon variants from Table~\ref{tab:optimizer-variants} and comparing them against AdamW baselines. Figure~\ref{fig:gpt2-medium-demo} plots validation loss versus training step. Under this setup, \texttt{AdamW-8L} diverges early (confirmed in separate runs and therefore omitted from the figure), consistent with the observations of \citet{dettmers20218}.

The AdamW baselines lag behind throughout training, with \texttt{AdamW-8D} consistently the worst among the plotted methods and \texttt{AdamW-32} also trailing the Muon family.
Secondly, all Muon-based variants---including hybrids that quantize the AdamW-associated states and variants that use either dynamic or linear quantization for Muon-associated states---track \texttt{Muon-32} extremely closely, with near-overlapping trajectories across the full training horizon.

\paragraph{Scaling results: pretraining validation loss.}
Having established stability and quantization robustness at the Medium scale, we next scale to a broader sweep over model sizes. We compare the full-precision \texttt{Muon-32} baseline against two practically relevant 8-bit variants, \texttt{Muon-8L/AdamW-32} and \texttt{Muon-8L/AdamW-8D}. We focus on these \texttt{Muon-8L}-based implementations because they require minimal engineering overhead, making them straightforward to integrate into distributed training frameworks (e.g., \cite{ahn2025dion}).

Table~\ref{tab:val_loss_multirow} reports the final pretraining validation loss for Small through XXL, aggregated over five seeds; the full training and validation curves are provided in Figures~\ref{fig:train_loss_pretrain} and~\ref{fig:val_loss_pretrain} in Appendix~\ref{appendix:train_val_loss}. Across all sizes, \texttt{Muon-8L/AdamW-32} matches full-precision \texttt{Muon-32} up to run-to-run variability: the mean loss differs by at most $0.002$ (e.g., Small: $3.120$ vs.\ $3.122$; XXL: $2.345$ vs.\ $2.346$). Meanwhile, Table~3 of \citet{dettmers20218} implies a loss gap of $0.006$ between 32-bit Adam and 8-bit Adam with block-wise dynamic quantization, both (i) with stable embeddings and (ii) without stable embeddings\footnote{We convert perplexity to loss via $\text{loss}=\ln(\text{Perplexity})$}.
\texttt{Muon-8L/AdamW-8D} remains competitive but shows a small and consistent degradation relative to \texttt{Muon-32}, with the largest gap at XL ($2.495$ vs.\ $2.508$). Overall, these results demonstrate that quantizing the Muon-associated state to 8-bit while keeping the AdamW component in full precision (\texttt{Muon-8L/AdamW-32}) preserves the optimization behavior of \texttt{Muon-32} across five model scales, while delivering the memory benefits of 8-bit state compression.

\paragraph{Downstream generalization}
We next assess generalization by evaluating each pretrained model on six established benchmarks. As shown in Table~\ref{tab:downstream_eval_results}, hybrids that keep the AdamW-associated states in full precision (e.g., \texttt{Muon-8L/AdamW-32}) remain on par with the full-precision \texttt{Muon-32} baseline across model sizes, with differences in the six-task average that are within seed-level variation. In contrast, the fully 8-bit hybrid \texttt{Muon-8L/AdamW-8D} exhibits a small but consistent degradation in the averaged downstream score, most noticeably at Medium, XL and XXL, mirroring its slightly worse pretraining validation loss. These results suggest that aggressively quantizing the AdamW component can be the limiting factor for downstream transfer, whereas quantizing the Muon-associated state alone preserves generalization.

\paragraph{Takeaway.}
Across a targeted ablation (Figure~\ref{fig:gpt2-medium-demo}) and a scale sweep (Tables~\ref{tab:val_loss_multirow}--\ref{tab:downstream_eval_results}), we find that (i)  Muon-based 8-bit variants train reliably; and (ii) \texttt{Muon-8L} matches full-precision \texttt{Muon-32} in both pretraining validation loss and downstream generalization across model sizes. These findings support \texttt{Muon-8L/AdamW-32} as a strong default for memory-efficient pretraining without sacrificing quality.


\begin{table}[t]
\centering
{\setlength{\tabcolsep}{5pt} 
\begin{tabular}{llc}
\toprule
\textbf{Model size} & \textbf{Optimizer} & \textbf{Validation loss} \\
\midrule
\multirow{3}{*}{Small}
 & \texttt{Muon-32}             & $\mstd{3.120}{0.002}$ \\
 & \texttt{Muon-8L/AdamW-32} & $\mstd{3.122}{0.002}$ \\
 & \texttt{Muon-8L/AdamW-8D} & $\mstd{3.129}{0.001}$ \\
\midrule
\multirow{3}{*}{Medium}
 & \texttt{Muon-32}             & $\mstd{2.819}{0.001}$ \\
 & \texttt{Muon-8L/AdamW-32} & $\mstd{2.820}{0.001}$ \\
 & \texttt{Muon-8L/AdamW-8D} & $\mstd{2.824}{0.001}$ \\
\midrule
\multirow{3}{*}{Large}
 & \texttt{Muon-32}             & $\mstd{2.620}{0.001}$ \\
 & \texttt{Muon-8L/AdamW-32} & $\mstd{2.621}{0.001}$ \\
 & \texttt{Muon-8L/AdamW-8D} & $\mstd{2.623}{0.001}$ \\
\midrule
\multirow{3}{*}{XL}
 & \texttt{Muon-32}             & $\mstd{2.495}{0.001}$ \\
 & \texttt{Muon-8L/AdamW-32} & $\mstd{2.496}{0.001}$ \\
 & \texttt{Muon-8L/AdamW-8D} & $\mstd{2.508}{0.001}$ \\
\midrule
\multirow{3}{*}{XXL}
 & \texttt{Muon-32}             & $\mstd{2.345}{0.001}$ \\
 & \texttt{Muon-8L/AdamW-32} & $\mstd{2.346}{0.001}$ \\
 & \texttt{Muon-8L/AdamW-8D} & $\mstd{2.352}{0.001}$ \\
\bottomrule
\end{tabular}}
\caption{Pretraining validation loss across 5 seeds, reported as mean with standard deviation.}
\label{tab:val_loss_multirow}
\end{table}

\begin{table*}[t]
\centering
\small
\setlength{\tabcolsep}{4pt}
\renewcommand{\arraystretch}{1.1}
\begin{tabular}{llccccccc}
\toprule
\textbf{Model size} & \textbf{Optimizer} & \textbf{ARC-C} & \textbf{ARC-E} & \textbf{BoolQ} & \textbf{HellaSwag} & \textbf{LAMBADA} & \textbf{MMLU} & \textbf{Avg. (6 tasks)} \\
\midrule
\multirow{3}{*}{Small} & \texttt{Muon-32} & $\mstd{0.204}{0.008}$ & $\mstd{0.503}{0.004}$ & $\mstd{0.571}{0.028}$ & $\mstd{0.285}{0.002}$ & $\mstd{0.203}{0.006}$ & $\mstd{0.230}{0.001}$ & 0.332 \\
 & \texttt{Muon-8L/AdamW-32} & $\mstd{0.211}{0.003}$ & $\mstd{0.505}{0.007}$ & $\mstd{0.574}{0.031}$ & $\mstd{0.285}{0.001}$ & $\mstd{0.202}{0.008}$ & $\mstd{0.230}{0.001}$ & \textbf{0.335} \\
 & \texttt{Muon-8L/AdamW-8D} & $\mstd{0.205}{0.009}$ & $\mstd{0.506}{0.008}$ & $\mstd{0.580}{0.023}$ & $\mstd{0.283}{0.001}$ & $\mstd{0.202}{0.006}$ & $\mstd{0.230}{0.000}$ & 0.334 \\
\midrule
\multirow{3}{*}{Medium} & \texttt{Muon-32} & $\mstd{0.236}{0.005}$ & $\mstd{0.564}{0.009}$ & $\mstd{0.606}{0.010}$ & $\mstd{0.316}{0.001}$ & $\mstd{0.272}{0.004}$ & $\mstd{0.231}{0.004}$ & 0.371 \\
 & \texttt{Muon-8L/AdamW-32} & $\mstd{0.247}{0.009}$ & $\mstd{0.563}{0.004}$ & $\mstd{0.613}{0.007}$ & $\mstd{0.315}{0.002}$ & $\mstd{0.274}{0.006}$ & $\mstd{0.232}{0.003}$ & \textbf{0.374} \\
 & \texttt{Muon-8L/AdamW-8D} & $\mstd{0.235}{0.008}$ & $\mstd{0.559}{0.010}$ & $\mstd{0.595}{0.011}$ & $\mstd{0.314}{0.002}$ & $\mstd{0.272}{0.006}$ & $\mstd{0.230}{0.002}$ & 0.368 \\
\midrule
\multirow{3}{*}{Large} & \texttt{Muon-32} & $\mstd{0.281}{0.008}$ & $\mstd{0.620}{0.014}$ & $\mstd{0.606}{0.012}$ & $\mstd{0.355}{0.002}$ & $\mstd{0.342}{0.005}$ & $\mstd{0.250}{0.006}$ & 0.409 \\
 & \texttt{Muon-8L/AdamW-32} & $\mstd{0.278}{0.006}$ & $\mstd{0.626}{0.007}$ & $\mstd{0.610}{0.010}$ & $\mstd{0.356}{0.002}$ & $\mstd{0.338}{0.004}$ & $\mstd{0.251}{0.009}$ & \textbf{0.410} \\
 & \texttt{Muon-8L/AdamW-8D} & $\mstd{0.272}{0.008}$ & $\mstd{0.628}{0.005}$ & $\mstd{0.611}{0.007}$ & $\mstd{0.355}{0.002}$ & $\mstd{0.335}{0.007}$ & $\mstd{0.245}{0.013}$ & 0.408 \\
\midrule
\multirow{3}{*}{XL} & \texttt{Muon-32} & $\mstd{0.296}{0.006}$ & $\mstd{0.653}{0.012}$ & $\mstd{0.607}{0.015}$ & $\mstd{0.382}{0.002}$ & $\mstd{0.376}{0.007}$ & $\mstd{0.255}{0.006}$ & \textbf{0.428} \\
 & \texttt{Muon-8L/AdamW-32} & $\mstd{0.298}{0.004}$ & $\mstd{0.651}{0.011}$ & $\mstd{0.603}{0.017}$ & $\mstd{0.381}{0.002}$ & $\mstd{0.373}{0.003}$ & $\mstd{0.253}{0.009}$ & 0.427 \\
 & \texttt{Muon-8L/AdamW-8D} & $\mstd{0.293}{0.005}$ & $\mstd{0.649}{0.007}$ & $\mstd{0.586}{0.037}$ & $\mstd{0.379}{0.001}$ & $\mstd{0.368}{0.004}$ & $\mstd{0.254}{0.006}$ & 0.422 \\
\midrule
\multirow{3}{*}{XXL} & \texttt{Muon-32} & $\mstd{0.354}{0.010}$ & $\mstd{0.713}{0.003}$ & $\mstd{0.618}{0.005}$ & $\mstd{0.428}{0.001}$ & $\mstd{0.424}{0.003}$ & $\mstd{0.261}{0.006}$ & \textbf{0.466} \\
 & \texttt{Muon-8L/AdamW-32} & $\mstd{0.352}{0.004}$ & $\mstd{0.707}{0.004}$ & $\mstd{0.618}{0.008}$ & $\mstd{0.428}{0.002}$ & $\mstd{0.422}{0.008}$ & $\mstd{0.256}{0.006}$ & 0.464 \\
 & \texttt{Muon-8L/AdamW-8D} & $\mstd{0.347}{0.008}$ & $\mstd{0.706}{0.013}$ & $\mstd{0.625}{0.009}$ & $\mstd{0.425}{0.002}$ & $\mstd{0.418}{0.005}$ & $\mstd{0.260}{0.010}$ & 0.463 \\
\bottomrule
\end{tabular}
\caption{Downstream performance across 5 seeds, reported as mean with standard deviation.}
\label{tab:downstream_eval_results}
\end{table*}


\begin{table}[t]
\vspace{-0.5em}
\centering
\small

\begin{tabular}{@{}l@{\hspace{6pt}}c@{\hspace{6pt}}c@{}}
\toprule
\textbf{Comparison} & \textbf{Win Rate (\%)} & \textbf{LC Win Rate (\%)} \\
\midrule
\texttt{Muon-32} vs \texttt{AdamW-32} & \textbf{59.67} & \textbf{59.17} \\
\texttt{M-8L/A-32} vs \texttt{AdamW-32} & \textbf{60.48} & \textbf{59.93} \\
\texttt{M-8L/A-32} vs \texttt{Muon-32} & 50.75 & 50.37 \\
\texttt{M-8L/A-8D} vs \texttt{AdamW-32} & 58.47 & 57.91 \\
\texttt{M-8L/A-8D} vs \texttt{Muon-32} & 48.17 & 48.14 \\
\bottomrule
\end{tabular}
\vspace{1.0em}
\caption{Head-to-head AlpacaEval results on $N{=}805$ prompts. Win Rate is the raw preference rate; LC Win Rate is length-controlled. Values are percentages; $50\%$ indicates a tie.}
\label{tab:alpacaeval-h2h}

\vspace{-1.0em}
\end{table}

\subsection{Fine-tuning with 8-bit Muon}

\textbf{Task Setup and Architecture}  We test the fine-tuning efficacy of the quantized Muon optimizer by using different variants to instruction-finetune the pre-trained models trained previously. Concretely, we choose two checkpoints of the GPT-XL model - one pre-trained with AdamW and the other pre-trained with Muon. It is empirically well-known that if a model is pre-trained with one optimizer, then empirically it is best to fine-tune it with the same optimizer~\cite{liu2025muon, team2025kimi}. Thus, we then proceed to finetune the AdamW-pretrained checkpoint with \texttt{AdamW-32}, and the Muon-pretrained checkpoint with \texttt{Muon-32}, \texttt{Muon-8L/AdamW-32} , and \texttt{Muon-8L/AdamW-8D}.

\textbf{Datasets} We use the Alpaca instruction following dataset~\cite{alpaca}. The dataset contains  about 52k samples for instruction following, consisting of a diverse array of topics from code generation to creative writing. 

\textbf{Training Details} We use SFT on the GPT-XL model using the learning rate of $2\times10^{-5}$. For each optimizer, 
For AdamW, we use $\beta_1=0.9$, $\beta_2=0.999$, and $\epsilon=10^{-8}$. We fixed the weight decay to be $0.01$.

\textbf{Evaluation Criteria} We use AlpacaEval 2.0~\cite{dubois2024length} (referred to hereinafter as AE2) as our main eval benchmark. AE2 is an automated evaluation framework that leverages a strong LLM-as-a-judge to compare the instruction following abilities of two different candidate LLMs on a variety of 805 prompts. AE2 reports both win rate and length-controlled (LC) win rate to mitigate verbosity bias. In our setup, we use GPT4 Turbo as the judge.

Once trained, we pit the four models against each other in a pairwise fashion to collect both win rate and LC win rate.

\subsubsection{Results}

Head-to-head AE2 results can be found in Table~\ref{tab:alpacaeval-h2h}. The model fine-tuned with \texttt{Muon-32} consistently outperforms the model fine-tuned with  \texttt{AdamW-32}, achieving 59.67\% win rate (59.17\% LC win rate). \texttt{Muon-8L/AdamW-32} performs similarly, achieving a large win over \texttt{AdamW-32} and being tied in performance with \texttt{Muon-32}. \texttt{Muon-8L/AdamW-8D} also outperforms \texttt{AdamW-32} and slightly underperforms when compared to \texttt{Muon-32}.

These findings mirror the results from pre-training, where we demonstrate that \textbf{quantizing the Muon-associated part of the optimizer results in essentially no loss in model quality}.



\subsection{Memory footprint}

Table~\ref{table:opt-state-mem} compares the persistent High Bandwidth Memory
 footprint of the optimizer state for variants, when profiled for the GPT XL and XXL models. For the XXL model, \texttt{Muon-8D/AdamW-32} provides substantial relative savings of \textbf{$\sim$79\%} and \textbf{$\sim$62\%} when compared to \texttt{AdamW-32} and  \texttt{Muon-32}. \texttt{Muon-8D/AdamW-8D} is more aggressive, stretching the advantage to \textbf{$\sim$86\%} and \textbf{$\sim$75\%}.

For the smaller XL model, the savings are slightly smaller. This is because the size of the embedding and lm-head matrices remains the same across model sizes because of a fixed vocabulary.

\begin{table*}[h!]
\centering
\sisetup{
  detect-weight,
  mode=text,
  round-mode=places,
  round-precision=2
}

\begin{tabular}{
    l
    S[table-format=2.2]
    S[table-format=2.2]
    S[table-format=2.2]
    S[table-format=2.2]
    >{\columncolor{bestcol}}S[table-format=2.2]
}
\toprule
Model Size & {\texttt{AdamW-32}} & {\texttt{Muon-32}} & {\texttt{AdamW-8D}} & {\texttt{M-8L(D)/A-32}} & {\texttt{M-8L(D)/A-8D}} \\
\midrule
XL  & 10.54 &  6.04 & 2.63 & 2.66 & \textbf{1.51} \\
XXL & 20.67 & 11.30 & 5.17 & 4.27 & \textbf{2.82} \\
\bottomrule
\end{tabular}
\caption{Optimizer states memory (GiB) for two model configurations (lower is better). \texttt{M-8L(D)/A-8D} yields the smallest optimizer-state footprint in both cases. }
\label{table:opt-state-mem}
\end{table*}
\section{Understanding the Stability of Linear Quantization}

Our experimental results in Section \ref{sxn:experiments_main} emphasize the effectiveness of applying linear quantization to the Muon optimizer state. This is in stark contrast to AdamW, where linear quantization results in divergence and even sophisticated dynamic quantization schemes may show performance degradation. In this section, we determine the mechanism driving these observations.

\subsection{Why is 8-bit AdamW Unstable under Linear Quantization?}\label{sxn:theory_main_text}

\citet{dettmers20218} observed that AdamW with naïve 8-bit \emph{linear} quantization performs poorly because it allocates too little resolution to small-magnitude entries, yielding large relative errors precisely where optimizer states concentrate most of their mass. We make this behavior more precise in Theorem \ref{thm:adam_quant} by analyzing how quantization error propagates and is magnified through the second-moment accumulator $\vb$. In particular, we show that if moderate-size gradient coordinates occur with non-negligible probability, then the expected squared error of one step of Adam with linearly quantized states \emph{diverges} as the numerical stabilizer $\epsilon \to 0$. The result holds for standard Adam hyperparameters; constants are kept explicit to emphasize practical regimes (e.g., $\epsilon \approx 10^{-8}$) where the error is already proved to be orders of magnitude larger than the unquantized update norm.

Let $Q$ denote the 8-bit linear quantization operator (Definition~\ref{def:8bit-linear}). Note that our error anlaysis is formalized over real numbers, and so $Q$ composes the quantization and dequantization steps that perturb the optimizer state. We consider the base Adam algorithm without weight decay, though it extends to AdamW without loss of generality\footnote{The result extends immediately to AdamW, since the decay of $\wb^{(0)}$ affects both $\tilde{\wb}^{(1)}$ and $\wb^{(1)}$ equally and therefore cancels out.}. All algorithmic details, definitions, and proofs are provided in Appendix~\ref{sxn:theory}. Note that we will use lower case letters when referring to one-dimensional vectorized parameters and upper case when operating on the two-dimensional hidden state parameters addressed by Muon, e.g., $\wb$ versus $\Wb$.

\begin{theorem}\label{thm:adam_quant}
    Let $\wb^{(1)}$ denote the parameters after one step of Adam 
    as given in Algorithm~\ref{alg:adam}, and let 
    $\tilde{\wb}^{(1)}$ denote the parameters after one step of the same algorithm 
    with 8-bit linear quantization applied to the moment estimates 
    (Definition~\ref{def:8bit-linear}), i.e.:
    \begin{gather*}
        \tilde{\wb}^{(1)} 
        = \wb^{(0)} - \alpha \cdot 
        \frac{Q(\mb^{(1)})}{\sqrt{Q(\vb^{(1)})} + \epsilon}.
    \end{gather*}
    Suppose that each entry of gradient $\gb^{(1)} \in \R^d$ satisfies 
    $\PP\!\left(\tfrac{\|\gb\|_\infty}{60} < |\gb_i| < \tfrac{\|\gb\|_\infty}{16}\right) \ge \nu$ 
    and $\|\gb\|_\infty \ge g_\infty > 256\,\epsilon$ with probability one. Then
    \begin{gather*}
       \mathbb{E}\,\|\wb^{(1)} - \tilde{\wb}^{(1)}\|_2^2 
       \;\ge\; \frac{d \nu\, \alpha^2 g_\infty^2}{(256\,\epsilon)^2}.
    \end{gather*}
\end{theorem}

We note that empirical validation of Theorem~\ref{thm:adam_quant}, as well as Theorems~\ref{thm:sgd_quant} and~\ref{thm:muon_quant_true_polar} below, can be found in Appendix~\ref{appendix:theorem_validation}.

\subsection{The Curious Case of 8-bit SGD with Momentum}

The proof of Theorem~\ref{thm:adam_quant} shows that the instability of Adam with linear quantization arises primarily from error in the second-moment vector, which appears in the denominator of the update rule. This naturally raises the question: \textbf{does 8-bit linear quantization suffice when such a denominator is avoided?} 

From a theoretical perspective, we show that, unlike Adam, SGD with momentum admits a uniform error bound under linear quantization. In particular, for any initialization of the weights and momentum, the quantization error remains bounded. Let $\eta > 0$ denote the step size and $\rho \in [0,1)$ the momentum parameter. 

\begin{theorem}\label{thm:sgd_quant}
    Consider a step of SGD with momentum with and without 8-bit linear quantization of the momentum:
    \begin{gather*}
        \tilde{\wb}^{(t+1)} = \wb^{(t)} - \eta(\gb^{(t)} + \rho Q(\mb^{(t)})) \\
        ~~\text{and}~~
        \wb^{(t+1)} = \wb^{(t)} - \eta(\gb^{(t)} + \rho \mb^{(t)}).
    \end{gather*}
    From any point $\wb^{(t)}$ and any momentum state $\mb^{(t)}$, then with linear quantization, $Q$ (Definition~\ref{def:8bit-linear}),
    \begin{gather*}
        \|\tilde{\wb}^{(t+1)} - \wb^{(t+1)}\|_2^2 \le d \eta^2\rho^2  \left(\frac{\|\mb^{(t)}\|_\infty}{127}\right)^2.
    \end{gather*}
\end{theorem}

Empirically, we confirm this difference. We train a ResNet-50 model~\citep{he2016deep} on the ImageNet dataset~\citep{deng2009imagenet}, using a standard training regime of 90 epochs. We compare AdamW and variants of SGD with momentum. Results are in Table~\ref{tab:sgd_vs_adam}. \textbf{We observe that SGD with linear 8-bit quantization achieves the same high validation top-1 accuracy~\citep{goyal2017accurate} of 76\%+ as full-precision SGD}. AdamW underperforms when compared to SGD (a well-known result on image classification training), while AdamW with linear quantization diverges immediately. See Appendix~\ref{appendix:imagenet} for extended details.

\begin{table}[t]
    \centering
    \small
    \begin{tabular}{@{}lcc@{}}
        \toprule
        \textbf{Method} & \textbf{SGD+M} & \textbf{AdamW} \\
        \midrule
        FP32                 & 76.21 & 74.42 \\
        8-bit linear quant. & 76.25 & --- \\
        \bottomrule
    \end{tabular}
    \caption{Top-1 validation accuracy (\%) after 90 epochs for SGD+M and AdamW in FP32 and with 8-bit linear quantization. 
    ``---'' indicates that AdamW with linear quantization diverged.}
    \label{tab:sgd_vs_adam}
\end{table}

Together, these results emphasize that quantized Adam’s instability is driven specifically by quantizing the second-moment term in the denominator for moderately small coordinates.

\subsection{Linear Quantization of the Muon Optimizer}

Like SGD with momentum, the Muon optimizer only maintains a first order momentum state between iterations, avoiding the second moment normalization term in AdamW. Theorem \ref{thm:muon_quant_true_polar} below shows that the Muon optimizer admits an analogous uniform error bound, theoretically explaining the stability of quantized Muon.

In Theorem~\ref{thm:muon_quant_true_polar} we show the following: when Muon uses an exact orthogonalization procedure (via the SVD), the quantization error bound for a single layer matches the SGD case up to an additional dependence on the smallest singular value, $s$, of the momentum matrix. This dependence is natural, since the conditioning of the momentum controls the stability of the orthogonal factor. 
Our analysis considers the exact polar factor to avoid introducing additional finite-iteration Newton–Schulz approximation error into the bound.

\begin{theorem}\label{thm:muon_quant_true_polar}
    Consider a step of Muon with momentum (using the exact polar factor rather than the Newton--Schulz approximation) with and without 8-bit linear quantization, $Q$ (Definition \ref{def:8bit-linear}), applied to the momentum. Let the layer weights and momentum state be $\Wb^{(t-1)}$ and $\Mb^{(t-1)}$, each with $d$ entries (see Appendix~\ref{sxn:exact_muon} for full update formulas).

    Suppose that, after a single gradient update, both the original and quantized momentum matrices are full column rank with minimum singular value at least $s>0$. Then
    \[
        \|\widetilde{\Wb}^{(t)} - \Wb^{(t)}\|_{\mathrm{F}}^2
        \;\leq\; \frac{d \alpha^2\beta^2}{s^2}
        \left(\frac{\|\vec(\Mb^{(t-1)})\|_\infty}{127}\right)^2,
    \]
    where $\widetilde{\Wb}^{(t)}$ denotes the weights after the quantized update and $\Wb^{(t)}$ after the unquantized update.
\end{theorem}

\section{Conclusions}

In this paper, we introduced 8-bit Muon, a memory-efficient variant of the Muon optimizer based on blockwise quantization of its optimizer state. We find that 8-bit Muon can match full-precision Muon in both validation loss and downstream benchmarks on Chinchilla-optimal GPT pretraining up to 2.7B parameters, while substantially reducing optimizer-state memory. In contrast to quantized AdamW, Muon remains compatible with simple linear quantization, providing a practical and easy-to-implement default. We complement these empirical results with a mechanistic and theoretical analysis that explains the differing behavior of AdamW, SGD with momentum, and Muon under quantization. Future work includes extending these methods to lower-bit quantization and combining them with complementary memory-saving techniques such as low rank momentum updates.

\bibliography{references}
\bibliographystyle{icml2026}

\onecolumn
\appendix
\section{Appendix}

\subsection{Training Hyperparameter Details}

\subsubsection{ImageNet training details}
\label{appendix:imagenet}

We trained ResNet-50 on ImageNet for 90 epochs using two H100 GPUs with PyTorch DistributedDataParallel. The schedule was the standard 90-epoch multi-step regime with learning rate decays at epochs 30, 60, and 80. For SGD with momentum we used a batch size of 128 per GPU (256 total), momentum $0.9$, and weight decay $10^{-4}$. For AdamW we used a learning rate of $3\times 10^{-3}$ and weight decay $10^{-2}$. Hyperparameters were held fixed across FP32 and quantized runs. Training images were augmented with random resized crops to $224\times224$ and random horizontal flips. At evaluation time, images were resized to 256 pixels on the short side and center-cropped to $224\times224$, followed by normalization with the standard ImageNet mean and variance.

In the quantized variants, optimizer states were stored in 8-bit linear form with per-tensor scaling (Definition~\ref{def:8bit-linear} applied layer-wise). For SGD with momentum, only the momentum buffer was quantized. For AdamW, both the first- and second-moment estimates were quantized. At each step, stored values were dequantized for computation, updated, and then requantized. Model weights, gradients, and activations were always maintained in FP32.

\begin{figure}[H]
    \centering
    \includegraphics[width=0.65\linewidth]{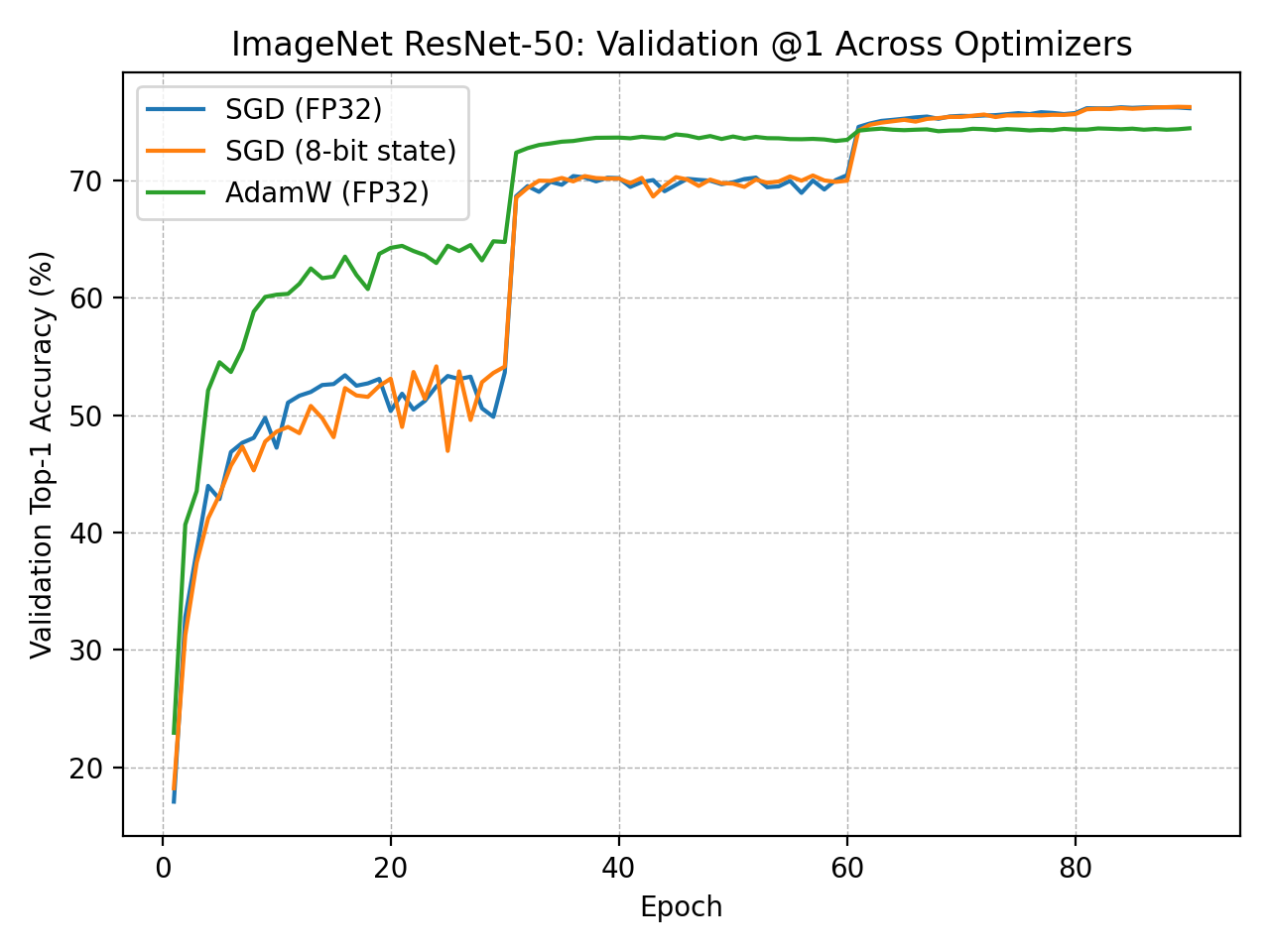}
    \caption{Validation top-1 accuracy on ImageNet for ResNet-50. Quantized SGD overlaps with the FP32 baseline. AdamW with FP32 states underperforms slightly, while the quantized AdamW variant diverged at the first step and is not shown.}
    \label{fig:imagenet_val1}
\end{figure}

Figure~\ref{fig:imagenet_val1} reports validation accuracy during training. Quantized SGD matches the FP32 baseline throughout. AdamW with FP32 optimizer states achieves slightly lower accuracy, while the quantized AdamW variant diverged immediately and is not shown.

\subsubsection{Pre-training Details}

\label{appendix:pretrain}

We train five GPT2-style models using 8–32 B200 GPUs with PyTorch Distributed DataParallel (DDP). Table~\ref{table:model_architectures} summarizes the model architectures, and Table~\ref{table:training_config} summarizes the pretraining hyperparameters.

\begin{table}[t]
    \centering
    \small
    \begin{tabular}{@{}lccccc@{}}
        \toprule
        \textbf{Model size} & \textbf{Parameters} & \textbf{$d_{model}$} & \textbf{$n_{layers}$} & \textbf{$n_{heads}$} & \textbf{FF Ratio} \\ 
        \midrule
        Small  & 162M   & 768  & 12 & 12 & 4 \\
        Medium & 405M   & 1024 & 24 & 16 & 4 \\
        Large  & 834M   & 1536 & 24 & 16 & 4 \\
        XL     & 1.4B   & 2048 & 24 & 16 & 4 \\
        XXL    & 2.7B   & 2560 & 32 & 32 & 4 \\ 
        \bottomrule
    \end{tabular}
    \caption{Model sizes used for pre-training. Parameter counts are slightly different than those reported by \citep{brown2020language} as the full architectural details were not reported. As a result, subtle differences in architecture as vocabulary size, GQA vs MHA and others may impact final model size.}
    \label{table:model_architectures}
\end{table}

\begin{table}[t]
    \centering
    \small
    \begin{tabular}{@{}lccccc@{}}
        \toprule
        \textbf{Model size} & \textbf{LR} & \textbf{Local BS} & \textbf{Grad. Acc. Steps} & \textbf{N. GPUs} & \textbf{Global BS} \\ 
        \midrule
        Small  & $6 \cdot 10^{-4}$ & 65536 & 1 & 8 &0.5M \\
        Medium & $3 \cdot 10^{-4}$ & 65536 & 1 & 8 &0.5M \\
        Large  & $2.5 \cdot 10^{-4}$ & 65536 & 1 & 8 & 0.5M \\
        XL     & $2 \cdot 10^{-4}$ & 32768 & 1 & 32 & 1M \\
        XXL    & $1.6 \cdot 10^{-4}$ & 32768 & 2 & 32 & 1M \\
        \bottomrule
    \end{tabular}
    \caption{Training configuration for the pre-training task. LR refers to the learning rate, and BS refers to the batch size. Both local and global batch sizes reported are in number of tokens, as all models were trained with a context length of 2048.}
    \label{table:training_config}
\end{table}

\section{Quantization Error Bounds}\label{sxn:theory}

In this appendix we formalize the 8-bit linear quantization operator and provide detailed proofs of the error bounds for Adam and SGD with momentum under quantization. We adopt the same notation as in Algorithm~\ref{alg:adam}: $\gb$ is the stochastic gradient, $\mb$ and $\vb$ are the first- and second-moment accumulators, $\gb^2$ denotes the entrywise square, $\sqrt{\vb}$ denotes the entrywise square root, and $\vec(\Mb)$ denotes the vectorization of $\Mb$. 

\subsection{Quantization Operator}

\begin{definition}\label{def:8bit-linear}
    (Linear quantization). For a given vector $\xb \in \R^d$ denoting the optimizer state of some algorithm, the 8-bit linear quantization is denoted by $Q:\R^d \rightarrow \R^d$, where:
    \begin{gather*}
        [Q(\xb)]_i = \frac{\|\xb\|_\infty}{127} \cdot \operatorname{round}\left(\frac{127 \cdot \xb_i}{\|\xb\|_\infty}\right).
    \end{gather*}
\end{definition}

That is, each coordinate of $\xb$ is mapped to the nearest grid point in a uniform partition of $[-\|\xb\|_\infty,\|\xb\|_\infty]$ into $256$ representable levels (corresponding to signed 8-bit integers from $-128$ to $127$), then rescaled back to floating point. This is the standard max-abs scaling scheme used in prior 8-bit quantization work. Note that, although we work over the reals, this definition effectively models the quantization and de-quantization steps applied to optimizer states between iterations.

\subsection{Proofs}

\subsubsection{Proof of Theorem \ref{thm:adam_quant}}
\begin{proof}

As stated in the theorem, we work at the first step ($t=1$) and suppress iterate superscripts for notational clarity. By the moment definitions in Algorithm~\ref{alg:adam}, we have $\mb=\gb$ and $\vb=\gb^{2}$ (entry-wise), so for each coordinate $i$ we have $\sqrt{\vb_i}=|\gb_i|$. We first lower bound the per-coordinate deviation $\left|\frac{\mb_i}{\sqrt{\vb_i}+\epsilon}-\frac{Q(\mb_i)}{\sqrt{Q(\vb_i)}+\epsilon}\right|$ and then sum over $i$.

\begin{align*}
    \PP\left(\left(\frac{\mb_i}{\sqrt{\vb_i} + \epsilon}  - \frac{Q(\mb_i)}{\sqrt{Q(\vb_i)} + \epsilon}\right)^2 \geq t^2 \right)
    &=
    \PP\left(\left|\frac{\mb_i}{\sqrt{\vb_i} + \epsilon}  - \frac{Q(\mb_i)}{\sqrt{Q(\vb_i)} + \epsilon}\right| \geq t \right) \\
     &\geq \PP\left(\frac{|\gb_i| - \|\gb\|_\infty/127}{\sqrt{Q(\vb_i)} + \epsilon} - \frac{|\gb_i|}{\sqrt{\vb_i} + \epsilon} \geq t ~\text{and}~
     \vb_i \geq Q(\vb_i) \right)
\end{align*}

Note that $\sqrt{\vb_i} = \sqrt{\gb_i^2} = |\gb_i|$ and $Q(\vb_i) = 0$ implies $\vb_i \geq Q(\vb_i)$. Hence, following from above,
\begin{align*}
      \PP\left(\left(\frac{\mb_i}{\sqrt{\vb_i} + \epsilon}  - \frac{Q(\mb_i)}{\sqrt{Q(\vb_i)} + \epsilon}\right)^2 \geq t^2 \right)
     &\geq \PP\left( \frac{|\gb_i| - \|\gb\|_\infty/127}{\epsilon}- \frac{|\gb_i|}{|\gb_i| + \epsilon} \geq t ~\text{and}~
      Q(\vb_i) = 0 \right)
      \\
      &\geq \PP\left( \frac{|\gb_i| - \|\gb\|_\infty/127}{\epsilon}- 1 \geq t ~\text{and}~
      Q(\vb_i) = 0 \right)
\end{align*}

Define the event
$E_i := \{ \tfrac{\|\gb\|_\infty}{60} < |\gb_i| \le \tfrac{\|\gb\|_\infty}{16} \}.$
By the assumption of the theorem, $\PP(E_i)\ge \nu$, and on $E_i$ we have both $Q(\vb_i)=0$ and
\begin{align*}
\frac{|\gb_i|-\|\gb\|_\infty/127}{\epsilon}
&\ge \frac{\|\gb\|_\infty/60 - \|\gb\|_\infty/127}{\epsilon} \\
&\geq \frac{\|\gb\|_\infty}{128\epsilon}.
\end{align*}

Since $\|\gb\|_\infty \ge g_\infty$ almost surely and $g_\infty \ge 256\epsilon$,
\begin{align*}
\frac{\|\gb\|_\infty}{128\epsilon} - 1 
&\ge \frac{g_\infty}{128\epsilon} - 1 \\
&\ge \frac{1}{2}\cdot \frac{g_\infty}{128\epsilon}
= \frac{g_\infty}{256\epsilon}.
\end{align*}
Set $\tau_0 := \frac{g_\infty}{256\epsilon}$. Then, for $t=\tau_0$,
\begin{align*}
\PP\left(
\left|\frac{\mb_i}{\sqrt{\vb_i}+\epsilon}-\frac{Q(\mb_i)}{\sqrt{Q(\vb_i)}+\epsilon}\right|
\ge t
\right)
&\ge \PP(E_i) \\
&\ge \nu.
\end{align*}
Therefore, using $\EE[X^2]\ge t^2\PP(X\ge t)$ for nonnegative $X$,
\begin{align*}
\EE\left[
\left(\frac{\mb_i}{\sqrt{\vb_i}+\epsilon}-\frac{Q(\mb_i)}{\sqrt{Q(\vb_i)}+\epsilon}\right)^2
\right]
&\ge \nu\,\tau_0^{2}
= \nu\frac{g_\infty^2}{(256\epsilon)^2}.
\end{align*}
Summing over $i=1,\dots,d$ and using
$\wb^{(1)}-\tilde{\wb}^{(1)}
= \alpha\!\left(\frac{\mb}{\sqrt{\vb}+\epsilon}-\frac{Q(\mb)}{\sqrt{Q(\vb)}+\epsilon}\right)$,
\begin{align*}
\EE\|\wb^{(1)}-\tilde{\wb}^{(1)}\|_2^2
&\ge \alpha^2 d \nu \frac{g_\infty^2}{(256\epsilon)^2}.
\end{align*}

\end{proof}

\subsubsection{Proof of Theorem \ref{thm:sgd_quant}}
\begin{proof}
    We compare the two updates and isolate the effect of quantizing the momentum. The gradient terms cancel, leaving
    \begin{align*}
        \tilde{\wb}^{(t+1)} - \wb^{(t+1)}
        &= -\eta(\gb^{(t)} + \rho Q(\mb^{(t)})) + \eta(\gb^{(t)} + \rho \mb^{(t)}) \\
        &= -\eta\rho\,(Q(\mb^{(t)}) - \mb^{(t)}).
    \end{align*}
    Taking squared norms and expanding coordinatewise yields
    \begin{align*}
        \|\tilde{\wb}^{(t+1)} - \wb^{(t+1)}\|_2^2
        &= \eta^2\rho^2 \sum_{i=1}^d \left(Q(\mb^{(t)})_i - \mb_i^{(t)}\right)^2.
    \end{align*}
    By Definition~\ref{def:8bit-linear}, each coordinate is perturbed by at most $\|\mb^{(t)}\|_\infty/127$, i.e., $|Q(\mb^{(t)})_i-\mb_i^{(t)}|\le \|\mb^{(t)}\|_\infty/127$. Applying this inside the sum gives
    \begin{align*}
        \|\tilde{\wb}^{(t+1)} - \wb^{(t+1)}\|_2^2
        &\le \eta^2\rho^2 \sum_{i=1}^d \left(\|\mb^{(t)}\|_\infty/127\right)^2 \\
        &= d\,\eta^2\rho^2\left(\|\mb^{(t)}\|_\infty/127\right)^2,
    \end{align*}
    which is the claimed bound.
\end{proof}

\subsubsection{Proof of Theorem \ref{thm:muon_quant_true_polar}}

\begin{proof}
From the two updates,
\[
  \widetilde{\Wb}^{(t)} - \Wb^{(t)} \;=\; -\alpha\big(\widetilde{\Ob}^{(t)}-\Ob^{(t)}\big)
  \quad\Rightarrow\quad
  \|\widetilde{\Wb}^{(t)} - \Wb^{(t)}\|_{\mathrm{F}}
  \;=\; \alpha\,\|\widetilde{\Ob}^{(t)}-\Ob^{(t)}\|_{\mathrm{F}}.
\]
The momentum matrices $\Mb^{(t)}$ and $\widetilde{\Mb}^{(t)}$ are assumed to be full column rank with
$\sigma_{\min}(\Mb^{(t)}),\sigma_{\min}(\widetilde{\Mb}^{(t)}) \ge s>0$.
For full-column-rank matrices, the (rectangular) polar-factor map satisfies
\[
  \|\widetilde{\Ob}^{(t)}-\Ob^{(t)}\|_{\mathrm{F}}
  \;\le\; \frac{2}{\sigma_{\min}(\Mb^{(t)})+\sigma_{\min}(\widetilde{\Mb}^{(t)})}
           \,\|\widetilde{\Mb}^{(t)}-\Mb^{(t)}\|_{\mathrm{F}}
  \;\le\; \frac{1}{s}\,\|\widetilde{\Mb}^{(t)}-\Mb^{(t)}\|_{\mathrm{F}},
\]
(see \citep[Thm.~VII.5.1(a)]{bhatia2013matrix} and its extension to full column
rank in \citep[Thm.~2]{li1995new}). Hence
\[
  \|\widetilde{\Wb}^{(t)} - \Wb^{(t)}\|_{\mathrm{F}}
  \;\le\; \alpha\,\frac{1}{s}\,\|\widetilde{\Mb}^{(t)}-\Mb^{(t)}\|_{\mathrm{F}}
  \;=\; \alpha\,\frac{\beta}{s}\,\|Q(\Mb^{(t-1)})-\Mb^{(t-1)}\|_{\mathrm{F}}.
\]
By Definition~\ref{def:8bit-linear},
each entry changes by at most $\|\vec(\Mb^{(t-1)})\|_\infty/127$, so
\(
  \|Q(\Mb^{(t-1)})-\Mb^{(t-1)}\|_{\mathrm{F}}^2
  \le d\big(\|\vec(\Mb^{(t-1)})\|_\infty/127\big)^2.
\)
Squaring both sides completes the proof.
\end{proof}

\subsection{Adam algorithm}

For completeness, we reproduce the base Adam algorithm below (Algorithm 1 in \cite{kingma2014adam}). 

\begin{algorithm}[h]
\caption{Adam (using $\wb, \gb, \mb, \vb$)}
\label{alg:adam}
\begin{algorithmic}[1]
\STATE \textbf{Input:} step size $\alpha$, decay rates $\beta_1,\beta_2\in[0,1)$, $\epsilon>0$
\STATE \textbf{Initialize:} $\wb^{(0)}$, $\tilde{\mb}^{(0)}=\mathbf{0}$, $\tilde{\vb}^{(0)}=\mathbf{0}$
\FOR{$t=1,2,\dots$}
  \STATE $\gb^{(t)} \gets \nabla f_t(\wb^{(t-1)})$
  \STATE $\tilde{\mb}^{(t)} \gets \beta_1 \tilde{\mb}^{(t-1)} + (1-\beta_1)\gb^{(t)}$
  \STATE $\tilde{\vb}^{(t)} \gets \beta_2 \tilde{\vb}^{(t-1)} + (1-\beta_2)\,(\gb^{(t)})^{2}$
  \STATE $\mb^{(t)} \gets \tilde{\mb}^{(t)}/(1-\beta_1^t)$
  \STATE $\vb^{(t)} \gets \tilde{\vb}^{(t)}/(1-\beta_2^t)$
  \STATE $\wb^{(t)} \gets \wb^{(t-1)} - \alpha\, \mb^{(t)}\big/(\sqrt{\vb^{(t)}} + \epsilon)$
\ENDFOR
\end{algorithmic}
\end{algorithm}

\subsection{Exact Muon optimizer}\label{sxn:exact_muon}

For completeness, we provide the update formula for the exact Muon algorithm (without weight decay) here. As opposed to eqn.~(\ref{eq:muon-1}), this formula uses the exact polar factor $\Ob = \Ub^{(t)}\Vb^{(t)\top}$

\begin{equation}
\begin{aligned}
\Mb^{(t)} &:= \beta\,\Mb^{(t-1)} + \Gb_t,\\
\Mb^{(t)} &= \Ub^{(t)}\Sb^{(t)}\Vb^{(t)\!\top}\ \text{(thin SVD)},\\
\Ob^{(t)} &:= \Ub^{(t)}\Vb^{(t)\!\top},\\
\Wb^{(t)} &= \Wb^{(t-1)} - \alpha\,\Ob^{(t)}.
\end{aligned}
\label{eq:muon-svd-update}
\end{equation}
    
In the quantized variant, only the previous momentum is quantized and then de-quantized, leading to the following updates where $Q$ is defined in Definition \ref{def:8bit-linear}:
\begin{equation}
\begin{aligned}
\widetilde{\Mb}^{(t)} &:= \beta\,Q\!\big(\Mb^{(t-1)}\big) + \Gb_t,\\
\widetilde{\Mb}^{(t)} &= \widetilde{\Ub}^{(t)}\widetilde{\Sb}^{(t)}\widetilde{\Vb}^{(t)\!\top}\ \text{(thin SVD)},\\
\widetilde{\Ob}^{(t)} &:= \widetilde{\Ub}^{(t)}\widetilde{\Vb}^{(t)\!\top},\\
\widetilde{\Wb}^{(t)} &= \Wb^{(t-1)} - \alpha\,\widetilde{\Ob}^{(t)}.
\end{aligned}
\label{eq:muon-svd-update-quant}
\end{equation}

\subsection{Empirical Validation}
\label{appendix:theorem_validation}

In this section, we provide empirical validation for the theorems of Section \ref{sxn:theory_main_text} using a simple experimental setup. We train a two-layer fully-connected network with one hidden layer of width 256 and ReLU activation, mapping \(28 \times 28\) pixel inputs to 10 logits. Inputs are normalized with the standard MNIST mean and variance \((0.1307, 0.3081)\), and we use the standard training split from \texttt{torchvision} with a batch size of 128 and cross-entropy loss.

\textbf{Theorem \ref{thm:adam_quant}} - We measure that approximately 29\% of gradient entries in the loss gradient, $\gb$, fall within the range $[\|\gb\|_\infty/ 60, \|\gb\|_\infty/16]$ from a Kaiming Uniform initialization of the model weights. This indicates that the gradient coordinate assumption of Theorem \ref{thm:adam_quant} holds in practice, and hence provides a realistic explanation for the divergence observed after a single step of Adam. 

\textbf{Theorem \ref{thm:sgd_quant}} - We measure the quantization error $\|\tilde{\wb}^{(t+1)} - \wb^{(t+1)}\|_2^2$ as a proportion of the error bound in Theorem \ref{thm:sgd_quant} for the SGD+M optimizer with learning rate $0.1$ and momentum $0.9$. For all $t$ over $1400$ training steps, the ratio of the true squared $\ell_2$ quantization error and the error bound stays between 0.06 and 0.085. By taking the square root of both sides in the error bound, we see that the true $\ell_2$-norm quantization error is accurately predicted by the theoretical bound up to a small constant factor.

\textbf{Theorem \ref{thm:muon_quant_true_polar}} - We measure the quantization error $\|\widetilde{\Wb}^{(t)} - \Wb^{(t)}\|_F^2$ of the first layer weight matrix for the Muon optimizer with $\alpha = 0.2$ and $\beta = 0.95$. We observe that the ratio of the true squared Frobenius-norm quantization error to the bound of Theorem \ref{thm:muon_quant_true_polar} stays between $2\times10^{-6}$ and $1.4 \times 10^{-5}$. The gap is explained by the fact that the proof of our theorem depends on the weak assumption that original and quantized matrices have full column rank $r$ with the $r$-th singular value of each matrix lower bounded by $s$. In fact, this condition is pessimistic, as the perturbation of the weight matrix is not aligned with the $r$-th singular vector. More complex assumptions on the spectrum of $\Wb$, paired with probabilistic assumptions on the quantization noise and a polar-factor perturbation analysis that accounts for additional spectral information (see \citet{li2003new}, for example), would produce a tighter bound. However, this analysis is outside the scope of this work. 

\subsection{Training and Validation Loss During Pretraining}
\label{appendix:train_val_loss}

\begin{figure*}[t]
    \centering
    \includegraphics[width=\linewidth]{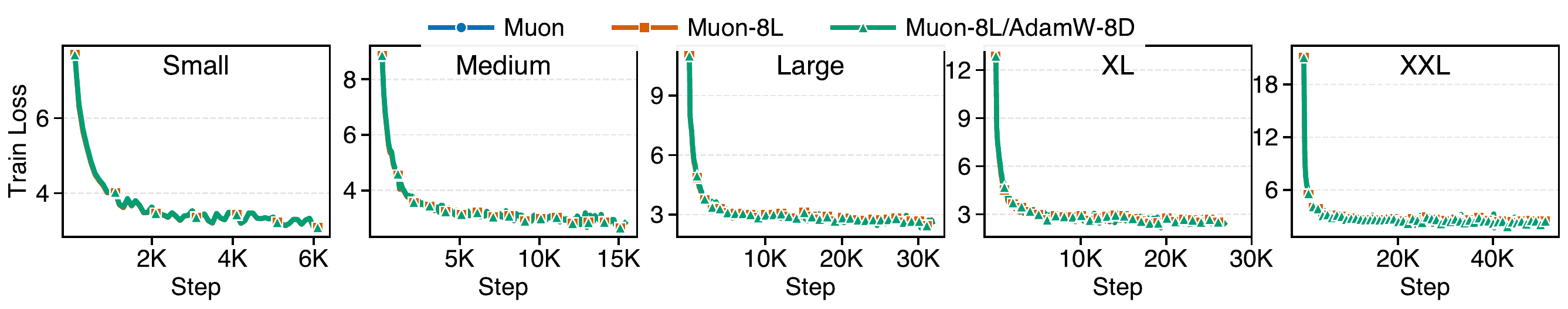}
    \caption{Pretraining training-loss curves for GPT-2 Small, Medium, Large, XL, and XXL with three optimizers: Muon, Muon-8L, and Muon-8L/AdamW-8D. Curves show the mean over 5 random seeds; error bars (seed-to-seed variation) are present but visually negligible.}
    \label{fig:train_loss_pretrain}
\end{figure*}

\begin{figure*}[t]
    \centering
    \includegraphics[width=\linewidth]{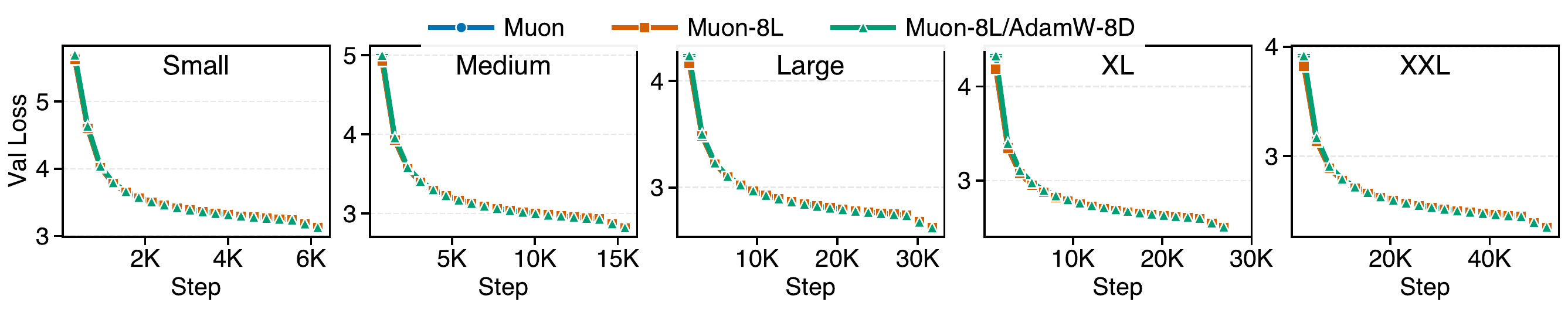}
    \caption{Pretraining validation-loss curves for GPT-2 Small, Medium, Large, XL, and XXL with three optimizers: Muon, Muon-8L, and Muon-8L/AdamW-8D. Curves show the mean over 5 random seeds; error bars (seed-to-seed variation) are present but visually negligible.}
    \label{fig:val_loss_pretrain}
\end{figure*}

\begin{figure*}[t]
    \centering
    \includegraphics[width=\linewidth]{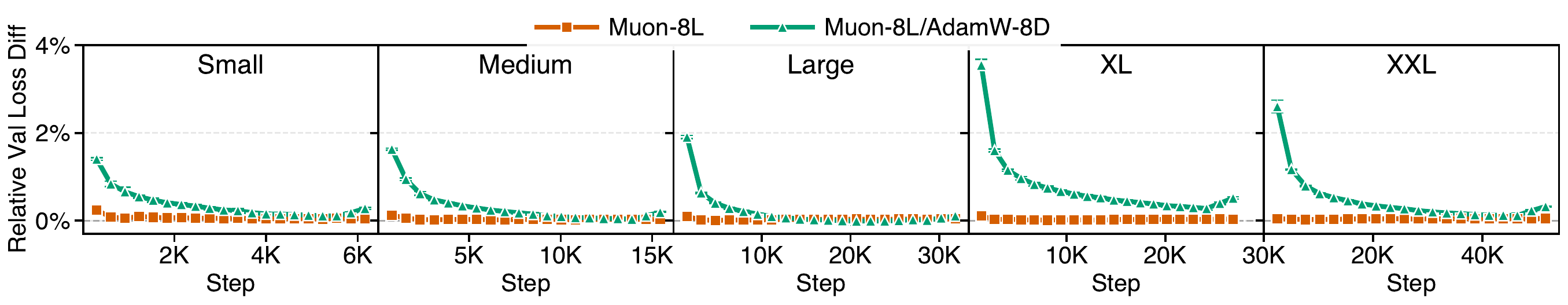}
    \caption{Relative pairwise validation-loss differences relative to the Muon baseline during pretraining for GPT-2 Small, Medium, Large, XL, and XXL.}
    \label{fig:val_loss_paired}
\end{figure*}

\end{document}